\documentclass[conference]{IEEEtran}
\IEEEoverridecommandlockouts 

\usepackage{amsmath}
\usepackage{amssymb}
\usepackage[nolist]{acronym}
\usepackage{bm}
\usepackage{color}
\usepackage{cite}
\usepackage[]{graphicx}
\usepackage[inline]{enumitem}
\usepackage{hyperref}

\graphicspath{{./figures}}

\newacro{dof}[DoF]{Degrees of Freedom}
\newacro{dl}[DL]{Deep Learning}
\newacro{ds}[DS]{Dynamical Systems}
\newacro{gmr}[GMR]{Gaussian Mixture Regression}
\newacro{ik}[IK]{Inverse Kinematics}
\newacro{il}[IL]{Imitation Learning}
\newacro{ilvs}[ILVS]{Imitation Learning Visual Servoing}
\newacro{nn}[NN]{artificial Neural Network}
\newacro{vs}[VS]{Visual Servoing}
\newacro{rds}[RDS]{Reshaped \acl{ds}}
\newacro{clfdm}[CLF-DM]{Control Lyapunov Function-based Dynamic Movements}
\newacro{clf}[CLF]{Control Lyapunov Function}
\newacro{fdm}[FDM]{Fast Diffeomorphic Matching}
\newacro{dmp}[DMP]{Dynamic Movement Primitive}
\newacro{seds}[SEDS]{Stable Estimator of Dynamical Systems}

\newcommand{\FriW}[1]{\emph{FriWalk}}

\begin{document}

\title{\LARGE \bf
Proactive Motion Planning for Human-Robot Cooperation
}

\author{Elena Basei$^{2,3}$, Edoardo Lamon$^{2,3}$, Matteo
  Saveriano$^{1,3}$, Daniele Fontanelli$^{1,3}$, Luigi
  Palopoli$^{2,3}$ \thanks{We acknowledge the support of the MUR PNRR
    project FAIR - Future AI Research (PE00000013), the project
    INVERSE (Grant Agreement n. 101136067), the project Magician
    (Grant Agreement n. 101120731) and the contribution of VRT
    foundation.}  \thanks{$^{1}$ Department of Industrial Engineering,
    Università di Trento, Trento, Italy.}  \thanks{$^{2}$
    Department of Information Engineering and Computer Science,
    Università di Trento, Trento, Italy.}  \thanks{$^{3}$
    Interdepartmental Robotics Labs (IDRA), University of Trento,
    Trento, Italy {\tt\small \{name.surname\}@unitn.it}} \thanks{The
    data collection and experiments were conducted with the approval
    of the ethics committee at the Authors' Institution under
    application N.2025-003.}  }

\maketitle
\thispagestyle{empty}
\pagestyle{empty}

\begin{abstract}
This abstract addresses the incorporation of human motion prediction into proactive and dynamic human-aware motion planning, with the goal of enabling safe collaboration between humans and robots. A deep learning, graph-based model is used to forecast human motion and is integrated into a planning framework. This framework employs a static roadmap along with a time-variant A* algorithm to modify the trajectory of a UR5e manipulator. This method greatly improves human-robot interaction and enables proactive collision avoidance by combining precise motion forecasts with adaptive trajectory planning.
\end{abstract}



\section{Introduction}\label{sec:intro}

Human-aware motion planning (HAMP) addresses the challenge of generating robot trajectories in shared workspaces while ensuring safety, efficiency, and explainability. Early solutions mainly relied on deliberative approaches~\cite{lavalle2006planning}, where robots computed complete paths in advance using graph-based or optimisation techniques. While effective in structured environments, these methods struggle in human-robot cooperation due to the dynamic nature of human presence, which can constantly alter the workspace. This often renders precomputed plans invalid and necessitates costly replanning. In response, reactive planning~\cite{MerckaertConstraint} emerged as an alternative by utilising real-time sensory feedback to adapt trajectories. Although this approach enhances safety, it treats humans as unpredictable obstacles, resulting in abrupt stops, jerky motions, and significant reductions in productivity.
Proactive planning ~\cite{liu2024integrating, HuppiTPRM} addresses these limitations by considering both current and predicted future human movements. By anticipating intent, robots can move more fluidly and safely, improving efficiency and operator comfort. Our work integrates motion forecasting into a human-aware planning framework, using a deep learning graph-based method to forecast human motion in real time and generate optimal, collision-free plans for robots.

The contributions are the following:
\begin{enumerate}[noitemsep,nolistsep]
    \item a real-time proactive planning method that utilises predicted human motion in conjunction with a time-variant A* algorithm;
    \item a parallelised version of the A* algorithm, optimised for real-time pathfinding in complex environments.
\end{enumerate}

\section{Methodology}\label{sec:method}

The proposed architecture for proactive motion planning includes three components: (i) online prediction of future human movements, (ii) an algorithm for generating collision-free robot trajectories, and (iii) a monitoring strategy that identifies potential collisions and replans as necessary. This work will focus on the last two components.

\subsection{Parallelised Time-Varying A*}\label{subsec:paral_A}
Our approach was inspired by the T-PRM framework by Hüppi et al.~\cite{HuppiTPRM} for planar holonomic robots. We extended the algorithm to address dynamic obstacles and to operate in real-time. First, we construct a graph $G = (V, E)$ from deterministic samples in 3D space based on end-effector positions, avoiding static obstacles. To optimise computation time, we reduce the configuration space from 6 DoFs to 3 by fixing the end-effector orientation. In the querying phase, we implement a modified A* algorithm that calculates the shortest, time-sensitive, collision-free paths. We introduce the arrival time map (\texttt{AT}($v$)), and the joint configuration map (\texttt{Cfg}($v$)) for each node. As we expand the nodes, a closed-form inverse kinematics (IK) solver is used to produce potential robot configurations, prioritising configurations based on distance from the previous one. Valid configurations must adhere to distance constraints and avoid collisions with the human position and predicted motion at the arrival time, in line with ISO/TS 15066. Invalid configurations are discarded. Once we identify a valid successor, we update \texttt{AT} and record the configuration in \texttt{Cfg}. After reaching the goal, we reconstruct the trajectory, resulting in a time-parametrised, collision-free motion plan $\pi_T(t)$ via polynomial spline interpolation.

To enhance computation efficiency, we developed a multithreaded A* algorithm. Instead of sequentially processing a queue of open nodes, our method makes use of $N$ threads that extract and evaluate nodes in a cycle while guaranteeing secure updates to shared resources.

\subsection{Plan Handling and Replan Strategy}\label{subsec:replan}
Although the multithreaded A* technique increases efficiency, it requires a significant amount of computing time, often more than inter-frame time. Moreover, unpredictable human movements can cause predictions to be inaccurate, leading to unrealistic plans or longer computation times. Lower speeds are also required by safety standards near critical limits. To address these issues, we use a state machine for execution oversight and parallel plan generation.

When a new data item arrives at time $T_i$, we start two processes: one generates a backup plan $\pi_{T_i}(t)$ and the other creates an emergency plan $\hat{\pi}_{T_i}(t)$ for quick recovery from unexpected system halts.
To generate $\pi_{T_i}(t)$, we run the modified A* algorithm using the HMP $h_{T_i}(t)$ with a deadline $D$ for plan delivery.  If the deadline is missed, no plan is produced. If met, the plan is valid and can be used starting from time $T_i + D$. The process begins with start position $s = FK\left(\pi_{\text{curr}}(T_i + D)\right)$, with $\pi_{\text{curr}}(t)$ denoting the current plan in execution. Since $D$ exceeds the inter-frame time $F$, multiple instances (up to $\left\lceil \frac{D}{F} \right\rceil$) can process different data items simultaneously. Once a plan $\pi_{T_i}(t)$ is successfully generated, it is stored for future use. 
If the robot needs to initiate motion, the starting position $s$ is set to the robot's current position.

The emergency plan $\hat{\pi}_{T_i}(t)$ is generated using the modified A* while keeping the human's position fixed and stopping once a thread reaches a position at time $F+D$. This approach reduces computation time, allowing us to generate $\hat{\pi}_{T_i}(t)$ within 0.01 seconds. The $F+D$ timing horizon ensures there is enough time for a new proactive plan to be developed.

During execution, when new data arrives at time $T_i$, a decision process evaluates if the current plan $\pi_{\text{curr}}(t)$ is still applicable based on the updated positions of the human. The process can switch between two states: \emph{Nominal Execution} and \emph{Collision Avoidance}. In the Nominal Execution state, the robot adjusts its speed based on its distance to the human, following ISO/TS 15066 guidelines. If the robot violates the stop distance or a collision is predicted at $T_i$, it transitions to Collision Avoidance, where for violations predicted after $T_i + D$, it schedules a transition to $\pi_{T_i}(t)$ at $T_i + D$. For violations before $T_i + D$, the robot immediately stops and sets the current plan to the emergency plan $\hat{\pi}_{T_i}(t)$. This re-planning approach ensures safe cooperation by allowing the robot to adapt its trajectory based on human actions.
\section{Results and Discussion}\label{sec:experiments}

\begin{figure}[t]
    \centering
    \includegraphics[width=0.7\columnwidth]{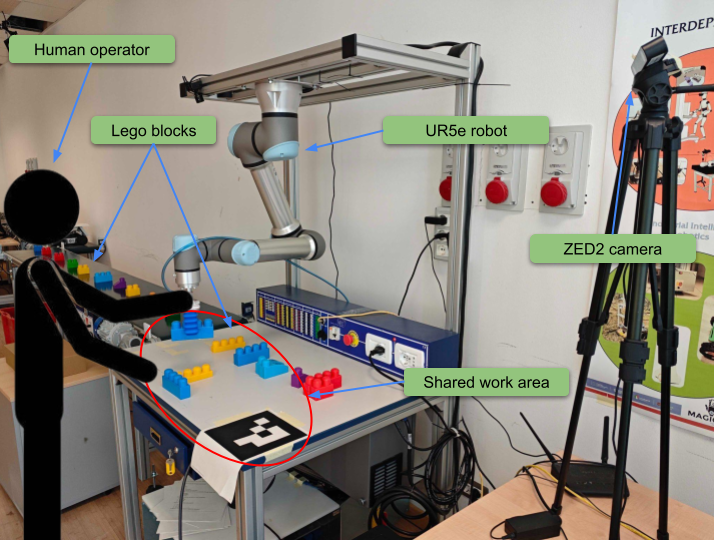}
    \caption{Test environment with Universal Robots UR5e and ZED2 camera.}
    \label{fig:test_env}
    \vspace{-5mm}
\end{figure}

We conducted a preliminary 5-minute evaluation of a cooperative assembly task using a Universal Robots UR5e and a ZED2 RGB-D camera to track human skeletons at 30 Hz. Ten subjects (4 females, 6 males, aged 23-31) with limited experience in robot manipulators performed the assembly task while the robot executed pick-and-place operations in the same area. Participants received blocks for assembly without specific instructions to test our replanning approach, and were advised to minimise movements in front of the robot to avoid triggering safety protocols. The setup included four drop stations and a conveyor belt pickup station (Fig.~\ref{fig:test_env}). We used the Graph-Mixer~\cite{YangAdaptive} network to predict human motion for the next second without fine-tuning it on our data, reflecting the typical constraints of small-medium enterprises where quick reconfiguration is often necessary. General applicability without task-specific fine-tuning is crucial in such scenarios.

We compared our proactive HAMP methods against a baseline provided by a reactive method, inspired by~\cite{MerckaertConstraint}, which only considers the current position of the human for replanning.
\begin{table}[t]
\centering
\caption{Comparison of proactive and reactive planning.}
\begin{tabular}{ c c c c c }
    Method & \begin{tabular}{@{}c@{}} Completed \\ Tasks\end{tabular} & \begin{tabular}{@{}c@{}}Avg. Planning \\ Time (s)\end{tabular} & \begin{tabular}{@{}c@{}} Avg. Stop \& \\ Re-plan\end{tabular} & \begin{tabular}{@{}c@{}} Avg. Online \\ Re-plan\end{tabular} \\
    \hline
    Reactive & $12.7$	& $0.18$ & $0.41$ & $0.49$ \\
    \textbf{Proactive} & $\textbf{15.5}$ & $\textbf{0.21}$ & $\textbf{0.20}$ & $\textbf{2.66}$ \\
\end{tabular}
\label{table:comparison_real}
\vspace{-5mm}
\end{table}
Table~\ref{table:comparison_real} displays the number of tasks completed, the average planning time, and the average number of stops and re-planning instances for each task. The initial results suggest that the proactive method outperforms the reactive approach. Specifically, the proactive approach showed a $22\%$ increase in task completion on average. Additionally, the average planning times for both methods appeared similar. Proactive methods also tended to reduce the number of stop-and-replan events by $53\%$. However, the proactive approach resulted in a higher online re-plan count, as it continuously updates the plan in the background, reflecting its anticipatory nature. Conversely, the reactive method exhibited non-zero re-plan counts ($0.49$), as it allowed for transitions between feasible plans whenever potential future collisions with the current human configuration were identified.

To obtain a preliminary indication of the relevance of these findings, we conducted a statistical analysis on the differences between the methods using either Wilcoxon signed-rank tests or repeated measures ANOVA, as appropriate, with a significance threshold at $p < 0.05$. These analyses suggested significant performance differences among the methods. Notably, there were differences in the completed tasks, with the proactive method tending to outperform the reactive baseline. Additionally, the mean online replanning test showed a highly significant result ($p = 1.15 \times 10^{-7}$), supporting the potential of the proactive approach. While these results are promising, further investigation is needed to draw definitive conclusions.



\bibliographystyle{IEEEtran}
\bibliography{bibliography}

\end{document}